\def\year{2019}\relax
\documentclass[letterpaper]{article} %DO NOT CHANGE THIS
\usepackage{aaai19}  %Required
\usepackage{times}  %Required
\usepackage{helvet}  %Required
\usepackage{courier}  %Required
\usepackage{url}  %Required
\usepackage{graphicx}  %Required
\usepackage{amsmath}
\usepackage{amssymb}

\newcommand{\citet}[1]{\citeauthor{#1} (\citeyear{#1})}
\newcommand{\citep}[1]{\cite{#1}}
\usepackage{float}
\usepackage{bibunits}

\newfloat{figtab}{htb}{fgtb}
\makeatletter
  \newcommand\tabcaption{\def\@captype{table}\caption}
\makeatother

\begin{document}

\title{Find a Reasonable Ending for Stories: \\ Does Logic Relation Help the Story Cloze Test?}
\author{Mingyue Shang\textsuperscript{1, $\diamond$}, Zhenxin Fu\textsuperscript{1}, Hongzhi Yin\textsuperscript{2}, Bo Tang\textsuperscript{4}, Dongyan Zhao\textsuperscript{1,3}, Rui Yan\textsuperscript{1,3}\thanks{Corresponding author: Rui Yan (ruiyan@pku.edu.cn) \newline Supported by the National Key R\&D Program of China (No.  2017YFC0804001) and the NSFC (No. 61672058\&61876196)}\\
\textsuperscript{1}{Institute of Computer Science and Technology, Peking University, China}~~
\textsuperscript{2}{The University of Queensland, Australia} \\
\textsuperscript{3}{Center for Data Science, Peking University, China}~~
\textsuperscript{4}{Southern University of Science and Technology, China} \\
\textsuperscript{$\diamond$}{Email: shangmy@pku.edu.cn}%, Phone: +86 13121966926}\\
}
\date{}

\maketitle
\begin{abstract}
Natural language understanding is a challenging problem that covers a wide range of tasks. While previous methods generally train each task separately, we consider combining the cross-task features to enhance the task performance. In this paper, we incorporate the logic information with the help of the Natural Language Inference (NLI) task to the Story Cloze Test (SCT). Previous work on SCT considered various semantic information, such as sentiment and topic, but lack the logic information between sentences which is an essential element of stories. Thus we propose to extract the logic information during the course of the story to improve the understanding of the whole story. The logic information is modeled with the help of the NLI task. Experimental results prove the strength of the logic information.
\end{abstract}

\section{Introduction}
Natural language understanding is an important field of Natural Language Processing which contains various tasks such as text classification, natural language inference (NLI) and story comprehension. These tasks differ in many aspects including the input formation and task goals. For example, the text classification task takes one sequence (a sentence or a document) as input to predict a label, while the NLI task takes two sequences as inputs to infer a relationship. Previous methods typically adopt supervised training to train a task-specific model. As the goal of tasks varies, the features that each model could extract also differs. While these features are treated separately in previous methods, it is intuitive that combining features across tasks may help to enhance the information representations. In this paper, we aim to explore the effectiveness of incorporating the features from one specific task to another task. 

Particularly, we investigate the feasibility of incorporating features learned in the NLI task into Story Cloze Test (SCT) \cite{mostafazadeh2016corpus} for the following reasons: datasets of these two tasks are in similar but still subtly different domains; the input and output form of the tasks are completely different which means distinct features are learned during the supervised training process. The NLI task predicts a relation from \textit{\{Entail, Neutral, conflict\}} of two sentences. The SCT is designed to choose a reasonable ending from two candidates, given four sentences as context (a.k.a plot). Compared with SCT, labels in the NLI task indicates explicit logic relations. In this paper, we explore to enhance the logic information representations with the help of NLI task to infer the relationship between the story context and the ending, as logic is an essential element of a story.

To this end, we propose a fully neural network based method that contains three components: a \emph{Content Unit} (CU) to encode the plot and the ending to get a content vector of the whole story; a \emph{Logic Unit} (LU)--the core unit to introduce logic features--to extract the logic relationship between each sentence in the plot and the ending, and then track the logic flow of the whole story; a \emph{Score Unit} (SU) to generate a score to indicate the reasonableness of the ending for the plot. Drawing upon modeling both content and logic into SCT, our method aims to provide new insights into story understanding task. 

\section{Architecture}
\begin{figure}[tbp]
  \centering
  \includegraphics[width = 0.45\textwidth]{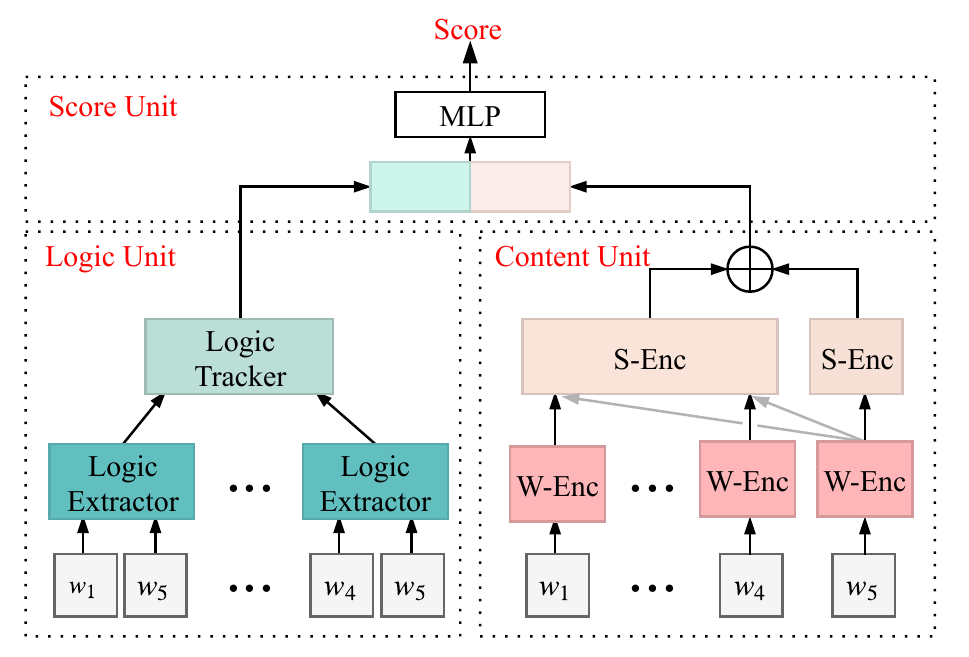}
  \caption{The overview of the proposed model which contains three main components.}
  \label{fig_mod}
\end{figure}

Figure \ref{fig_mod} gives an overview of the proposed model. Concretely, a plot $p = \{w_1, \ldots, w_4\}$ and a candidate ending $e = w_5$ are fed to both CU and LU, where $w_i$ means the $i$-th sentence. After that, the SU takes the output of CU and LU as input and produces a score. A higher score means the ending is more reasonable to the plot, and vise versa.

\noindent \textbf{Content Unit. }
Inspired by \citet{cai2017pay}, we implement the hierarchical recurrent neural network (RNN) to encode the 4-sentence plot and the ending into a content vector. %Note that the plot and the ending are encoded by the same RNNs. 
The hierarchical RNN contains a word level RNN (referred as W-Enc) and a sentence level RNN (referred as S-Enc), which are bi-directional GRU for both levels.

The W-Enc first converts each sentence $w_i$ into hidden states $h_{w, i} = \{h^i_{w, i}, \ldots, h^n_{w, i}\}$, and then computes the attention of the ending over each sentence through hidden states, denoted as $\beta_i = \{\beta^1_i, \ldots, \beta^n_i\}$. The sentence vector $s_i$ is calculated as the attention-weighted sum of the hidden states. Then the S-Enc encodes the plot $P = \{s_1, \ldots, s_4\}$ and the ending $E = \{s_5\}$, respectively. The final content vector $V_c$ is defined as $V_c = [u_p; u_e]$, where $u_p$ and $u_e$ are the hidden state of the plot $P$ and the ending $E$ at the final step, and $[;]$ means concatenation.

\noindent \textbf{Logic Unit. }
Logic Unit consists of a logic extractor to capture the logic relation of two sentences and a logic tracker to track the sequential logic information of the story. 

\textbf{Logic extractor.}~For the purpose of enhancing logic information purpose, we pre-train an NLI model on NLI dataset. The NLI model follows the structure of ESIM \cite{chen2017enhanced}. Readers can refer to \citet{chen2017enhanced} for more details. 
During the pre-training process, the ESIM takes two sentences as inputs and encodes them to vector representation, which is then fed to a multi-layer perceptron (MLP) with softmax operation to predict a label. The parameters are updated by cross-entropy loss. In this work, we discard the final layer of the MLP in the ESIM and employ the remainder as the extractor. When trained on the SCT, each sentence in the plot is coupled with the ending to form a sentence pair $p_i$. Given the pair as input, the extractor gives an intermediate vector $u_{l, i}$ which represents the logic information.\footnote{As the model is trained by the labels that requires the model to extract the logic relation, the output before the softmax layer in the MLP is supposed to contain the logic information as to correctly predict the label.} 

\textbf{Logic Tracker.}~Build upon the extractor, the logic tracker aims to model the logic flows of the whole story. 
Four sentence pairs are fed to the extractor respectively, and the outputs of the extractor form a sequence $L = \{u_{l, 1},\ldots, u_{l,4}\}$. The tracker is implemented by a BiGRU to encode $L$ into hidden states $h_t = \{h_{t,1}, \ldots, h_{t, 4}\}$. Then the sequential hidden states are concatenated as the final output of the logic tracker $V_l = [h_{t,1}; \ldots; h_{t, 4}]$.

\noindent \textbf{Score Unit. }
The final prediction is based on the combination of content information $V_c$ and the logic information $V_l$. They are concatenated and fed into a score function to predict a score $S$ which measures how reasonable the ending is. The score function $S$ is implemented by a MLP with $\tanh$ activation function.

As each plot is accompanied with a right ending and a wrong ending, we apply hinge loss to train the model. Concretely, $S^+$ denotes score of the story with the right ending and $S^-$ denotes the score with the wrong ending. The objective is to minimize $L$, where $L= max(0,~\Delta - S^+ + S^-)$,
$\Delta$ is the margin and is set to 1 in this paper.%, $\Delta$ is set to 1.

\begin{figtab}

\begin{minipage}[d]{0.5\linewidth}
\centering
\resizebox{0.77\textwidth}{12mm}{
  \begin{tabular}{|l|c|}
  \hline
  Model       & ACC.     \\ \hline
  HEPA        & 74.7\%   \\ \hline
  LS-Skip     & 76.5\%   \\ \hline
  HCM         & 77.6\%   \\ \hline
  FTLM        & \textbf{86.5}\%  \\ \hline
  Our Model   & 79.1\%   \\ \hline
  \end{tabular}
  }
  \tabcaption{Test-set results.} % of  \\ various models. }
  \label{tab_res}
  \end{minipage} 
  \begin{minipage}[d]{0.42\linewidth}
  \centering
  \resizebox{0.78\textwidth}{10mm}{
\begin{tabular}{|l|c|}
  \hline
  Model                                               & ACC. \\ \hline
  CU                                        & 71.7\%     \\ \hline
  LU                                          & 75.6\%     \\ \hline
  All                       & 79.1\%   \\ \hline
  \end{tabular}
  }
  \tabcaption{Ablation study.}% results.}
  \label{ablation_res}
  \end{minipage}
\end{figtab}
\section{Experiments}
\noindent \textbf{Dataset.} We conduct the story cloze experiments on the ROCStories corpus \citep{mostafazadeh2016corpus} which has 98,161 stories in the training set and 1,871 4-sentence cases in the validation set and test set respectively. Each 4-sentence case has a right ending and a wrong ending. The ESIM mentioned in the logic extractor is trained on an NLI dataset. Further details are provided in the supplement.%\footnote{https://shangmy.github.io/publication/story-cloze}.

\noindent \textbf{Experimental Results.} Table \ref{tab_res} shows the results of our approaches and other models . 
Among the reported baselines, HEPA \citep{cai2017pay} and LS-Skip \citep{srinivasan2018simple} are fully neural network based, but their performances are slightly weaker then others. The Hidden Coherence Model (HCM) \citep{chaturvedi2017story} relies on the feature engineering. Though our performance is weaker than the current state-of-the art result achieved by the FTLM \citep{radford2018improving}, compared with FTLM which requires large-scale pretrain, our model involves less data and focuses on investigating the use of logic information. The ablation results further verify the effectiveness of the cross-task information in Table \ref{ablation_res}. 

\section{Conclusion and Future Work}
In this paper, we aim to investigate the feasibility of incorporating cross-task information. We conduct experiments on SCT with the enhancement from the NLI task. In the future, we plan to explore the possibility of inter-task combination from other aspects.

\bibliographystyle{aaai}
\bibliography{aaai}

\section{Supplementary Materials}
\subsection{NLI Dataset}
We blend the SNLI corpus presented by \citet{bowman2015large} and the Multi-NLI corpus presented by \citet{williams2017broad} to form a new NLI dataset. Considering that most sentences in ROCStories are relatively short, we only keep the sentences in the blended corpus whose lengths are less than 20. We finally get 678,225 cases as training set and 1,000 cases as validation set. 

\subsection{Training} 
We first pre-train the ESIM and choose the model according to the performance on the NLI validation set. Then we train the whole model on validation set of SCT. The parameters of the extractor are tuned during that process. Concretely, we randomly split the validation set into 5 folds and use 5-fold cross validation. We adopt the pre-trained 300-dimensional GloVe \citep{pennington2014glove} vectors to initialize our word embeddings and use the Adam \citep{kingma2014adam} for optimization. The initial learning rate is set to 0.001 and the batch size is 32. In LU, the hidden size of GRU layers is 400 for the extractor and 128 for the tracker. The hidden size of all the GRU layers in CU is 512.

\begin{table}[!h]
  \centering
  \begin{tabular}{|p{7.2cm}|}
    \hline
    \textbf{Context}:                          \\
    Daddy took us to the woods to camp.                          \\ 
    He taught us how to build a fire with sticks.               \\
    He helped us learn how to make our own tents.              \\
    We fell asleep counting the stars in the sky.                \\ \hline
    \textbf{Right Ending}: \\
    We had a wonderful time camping.                      \\
    \textbf{Wrong Ending}: \\
    We will never camp again.    \\ \hline
  \end{tabular}
  \caption{An example from the validation set of SCT task.}  \label{cloze case}
\end{table}

\subsection{Cases}
Table \ref{cloze case} shows an example in SCT. It can be seen that the wrong ending is still closely relevant to the context, which means shallow language representations are of limited help. As mentioned in \citet{mostafazadeh2016corpus}, correctly understanding these stories requires richer semantic representations of events.

\end{document}